\documentclass[a4paper]{article}
\usepackage{times}
\usepackage{helvet}
\usepackage{courier}
\usepackage{url}
\usepackage{paralist}
\usepackage{multicol}
\usepackage{multirow}
\usepackage{algorithm}
\usepackage{algorithmic}
\usepackage{graphicx}
\usepackage{pstricks}
\usepackage{epsfig}
\usepackage{amsmath}
\usepackage{amssymb}

\newtheorem{example}{Example}

\begin{document}
%
\title{Distributional Framework for Emergent Knowledge Acquisition and its
Application to Automated Document Annotation}
\author{
V\'it Nov\'a\v{c}ek\\
Digital Enterprise Research Institute (DERI), \\
National University of Galway  Ireland (NUIG)\\
IDA Business Park, Lower Dangan\\
Galway, Ireland\\
e-mail: 
\url{vit.novacek@deri.org}\\
}
\maketitle
\begin{abstract}
The paper introduces a framework for representation and acquisition of 
knowledge emerging from large samples of textual data. We utilise a 
tensor-based, distributional representation of simple statements extracted 
from text, and show how one can use the representation to infer emergent 
knowledge patterns from the textual data in an unsupervised manner. Examples 
of the patterns we investigate in the paper are implicit term relationships or 
conjunctive IF-THEN rules. To evaluate the practical relevance of our 
approach, we apply it to annotation of life science articles with terms from 
MeSH (a controlled biomedical vocabulary and thesaurus). 
\end{abstract}


\section{Introduction}\label{sec:intro}

The ubiquity of methods for digital content publishing, processing and sharing 
has led to a lot of data being made globally available every day. Such an
unprecedented world-wide availability of content is generally beneficial, yet 
it also poses big challenges. For instance, in as dynamic and voluminous 
domains as life sciences, it is virtually impossible for the users to utilise
all the available relevant knowledge in a comprehensive and timely 
manner~\cite{tsujii2009}. 

Mitigation of this problem (with a special focus on
biomedical literature) has served as the main motivation for the research 
presented in this paper. As can be seen for instance 
in~\cite{huangetal2011jamia}, a popular way of tackling the information 
overload in the context of biomedical literature is annotation of articles by 
terms from standardised biomedical vocabularies. Such annotations can in turn 
make the retrieval of relevant documents much more efficient. However, as 
providing the necessary annotations manually is very expensive, automated 
methods are desired~\cite{huangetal2011jamia}, which is what we are going to 
address here. 

The technical contribution of the presented work is two-fold. Firstly, we 
introduce a general framework for automated acquisition of knowledge from 
textual collections. The proposed framework builds on the principles of 
distributional~\cite{dist_hyp1957} and emergent~\cite{cudre-mauroux2009emsem} 
semantics, and allows for inference of complex knowledge patterns 
within simple 
co-occurrence statements extracted from articles. As a second contribution, we 
show how the knowledge 
inferred from the text can be applied to 
unsupervised and parameter-free annotation of biomedical articles. 

The rest of the paper is organised as follows. Section~\ref{sec:related} 
gives an overview of related work. 
The framework for emergent knowledge representation and acquisition is 
described in Section~\ref{sec:theory}. The application of the framework to 
document annotation is detailed in Section~\ref{sec:practice}, where we also 
discuss an experiment we performed to evaluate our approach. 
Section~\ref{sec:conclusion} concludes the paper and outlines our future work.


\section{Related Work}\label{sec:related}


Our approach builds on and shares a lot of similarities with recent works in
emergent~\cite{cudre-mauroux2009emsem} and distributional~\cite{dm2010} 
semantics. However, \cite{cudre-mauroux2009emsem} is quite restrictive and
applies the notion of emergence~\cite{sw_emergence} merely to complex patterns
arising from simple interactions of autonomous agents in distributed systems 
like P2P networks. We 
are more general, focusing rather on inference and analysis of complex 
patterns emerging within large amounts of simple statements being extracted 
directly from data. This is in accordance with a recent approach to 
distributional semantics presented in~\cite{dm2010}. We employ similar 
tensor-based structures for representation of data and analysis of the 
knowledge emerging from them. Yet we also augment the work~\cite{dm2010} 
by an explicit representation of data provenance and, more importantly, by a 
method for mining rules out of the distributional representations. The latter 
is related to the associative rule mining introduced 
in~\cite{Agrawal:1993:MAR:170036.170072}, however, we generalise the state of 
the art method to make use of our distributional (essentially vector-based) 
representation of the data. 

Regarding the application of our framework to annotation of biomedical 
articles, a body of more or less recent works like~\cite{kimetal2001amia}, 
\cite{aaronsonetal2004hinf}, \cite{neveol2009jbi}, \cite{ruch2006bioinf} 
or~\cite{huangetal2011jamia} exists (the second, third and fifth of the 
approaches are either used or considered for use as a support service for the 
professional annotators of the articles on PubMed, a biomedical literature 
repository). The state of the art methods, however, often require at least an
indirect input from human users before they can produce annotations of new 
articles automatically. For instance, \cite{aaronsonetal2004hinf} and 
\cite{huangetal2011jamia} require a large corpus of previously annotated 
articles for learning and ranking possible annotations of new resources. 
Other methods like~\cite{neveol2009jbi} require rather sophisticated tuning 
(e.g., experimenting with parameter settings or with the processing pipeline 
composition) for optimum performance on new data. This is not the case of our 
approach, as it can work in a purely unsupervised manner off-the-shelf. 


\section{Distributional Framework for Emergent Knowledge 
Acquisition}\label{sec:theory}


This section first describes how one can represent the knowledge emerging from 
textual documents at various levels of complexity:
\begin{inparaenum}[(1)]
  \item simple term co-occurrence statements within the documents;
  \item an integral view on the statements across the document corpus;
  \item different perspectives of the corpus-wide view for analysing various 
  types of emergent semantic phenomena.
\end{inparaenum}
All levels of the representation are based on compact tensor structures 
(tensor is a generalisation of the scalar, vector and matrix notions; see 
\url{http://en.wikipedia.org/wiki/Tensor} for a more detailed overview).
The rest of the section deals with analysis of two particular types of emergent
semantic phenomena that are relevant to document annotation, our motivating 
use case. 


\subsection{Source Representation}
The first layer consists of a so called source representation $\mathbf{G}$, 
which captures the co-occurrence of terms across a set of documents (a 
concrete way of extracting co-occurrence relationships is presented in 
Section~\ref{sec:practice}). Let 
$A_l, A_r$ be sets representing left and right arguments of binary 
co-occurrence relationships (i.e., statements), and $L$ the types of the 
relationships. Furthermore, let $P$ be a set representing provenances of 
particular relationships (i.e., document identifiers). We define the source 
representation as a 4-ary labeled tensor $\mathbf{G} \in \mathbb{R}^{|A_l| 
\times |L| \times |A_r| \times |P|}$. It is a four-dimensional array structure 
indexed by argument-link-argument-provenance tuples, with values reflecting the
weight (e.g., frequency) of statements in the context of particular sources 
($0$ if a statement does not occur in a source). For instance, if a statement 
{\it (protein, different from, gene)} occurs two times in a source $d_x$, 
then the element $g_{protein,different\;from,gene,d_x}$ of $\mathbf{G}$ will 
be $2$. More details are given in Example~\ref{ex:source}.

\begin{example}\label{ex:source}
\small
Let us consider documents $d_1, d_2, d_3, d_4$ and the following terms 
occurring in them: \emph{protein domain, protein, domain, gene, internal 
tandem duplications, mutations, juxtamembrane, extracellular domains} 
(abbreviated as 
{\it p.d., p., d., g., i.t.d., m., j., e.d.}, respectively, in the following). 
Let us further assume that the following statements were extracted from the 
documents: $d_1: \{$ {\it (p.d., D, p.), (p.d., T, d.)} 
$\}$, $d_2: \{$ {\it (g., D, p.)} $\}$, $d_3: \{$ {\it (i.t.d., T, m.), 
(i.t.d., I, j.), (i.t.d., I, e.d.)} $\}$, $d_4: \{$ {\it (p.d., D, p.)} $\}$,
where {\it D, T, I} are abbreviations for relation terms \emph{different 
from, type of, in}. When omitting all zero values and representing a 
four-dimensional tensor as a two-dimensional table where the three first 
columns are for the tensor indices and the fourth one is for the corresponding 
tensor value, we can represent the source with the above statements as 
follows (using statement frequencies as values):
\begin{center}
\tiny
\begin{tabular}{|c|c|c|c||c|}
\hline
$s \in A_l$ & $p \in L$ & $o \in A_r$ & $d \in P$ & $g_{s,p,o,d}$ \\
\hline\hline
{\it p.d.} & {\it D} & {\it p.} & $d_1$ & $1$ \\ 
\hline
{\it p.d.} & {\it T} & {\it d.} & $d_1$ & $1$ \\ 
\hline
{\it g.} & {\it D} & {\it p.} & $d_2$ & $1$\\ 
\hline
{\it i.t.d.} & {\it T} & {\it m.} & $d_3$ & $1$ \\ 
\hline
{\it i.t.d.} & {\it I} & {\it j.} & $d_3$ & $1$ \\ 
\hline
{\it i.t.d.} & {\it I} & {\it e.d.} & $d_3$ & $1$ \\
\hline
{\it p.d.} & {\it T} & {\it d.} & $d_4$ & $1$ \\ 
\hline
\end{tabular}
\end{center}
\end{example}

\subsection{Corpus Representation}
The source tensor is a low-level data representation merely preserving the 
association of statements with their provenance contexts. Before allowing for 
actual distributional analysis, the data have to be transformed into a more 
compact structure $\mathbf{C}$ we call corpus representation. $\mathbf{C} \in 
\mathbb{R}^{|A_l| \times |L| \times |A_r|}$ is a ternary (three-dimensional) 
labeled tensor providing for a universal and compact distributional 
representation of simple statements extracted from source documents. A corpus 
$\mathbf{C}$ can be constructed from a source representation $\mathbf{G}$ 
using functions $a: 
\mathbb{R}\times\mathbb{R} \rightarrow \mathbb{R}, w: P \rightarrow 
\mathbb{R}, f: A_l \times L \times A_r \rightarrow \mathbb{R}$. For each 
$\mathbf{C}$ element $c_{s,p,o}$, $c_{s,p,o} = a(\sum_{d \in P}w(d)g_{s,p,o,d},
h(s,p,o))$, where $g_{s,p,o,d}$ is an element of the source tensor 
$\mathbf{G}$ and the $a, f, w$ functions act as follows:
\begin{inparaenum}[(1)]
  \item $w$ assigns a relevance degree to each document $d \in P$;
  \item $f$ reflects the relevance of the statement elements (e.g., mutual 
  information score of the subject and object within the source);
  \item $a$ aggregates the result of the $w,f$ functions' application.
\end{inparaenum}
This way of constructing the elements of the corpus tensor from the source 
representation aggregates the occurrences of statements within the input data, 
reflecting also two important things -- the relevance of particular sources 
(via the $w$ function), and the relevance of the statements themselves (via 
the $f$ function). The specific implementation of the functions is left to 
applications -- alternatives include (but are not limited to) ranking 
(both at the statement and document level) or statistical analysis of the 
statements within the input data. 

\begin{example}\label{ex:corpus}
\small
A corpus corresponding to the source tensor from Example~\ref{ex:source} can 
be represented as depicted below. The $w$ values were $1$ for all sources and 
$a,f$ aggregated the source values using relative frequency (in a data set 
containing $7$ statements in total).
\begin{center}
\tiny
\begin{tabular}{|c|c|c||c|}
\hline
$s \in A_l$ & $p \in L$ & $o \in A_r$ & $c_{s,p,o}$ \\
\hline\hline
{\it p.d.} & {\it D} & {\it p.} & $1/7$ \\ 
\hline
{\it p.d.} & {\it T} & {\it d.} & $2/7$ \\ 
\hline
{\it g.} & {\it D} & {\it p.} & $1/7$\\ 
\hline
{\it i.t.d.} & {\it T} & {\it m.} & $1/7$ \\ 
\hline
{\it i.t.d.} & {\it I} & {\it j.} & $1/7$ \\ 
\hline
{\it i.t.d.} & {\it I} & {\it e.d.} & $1/7$ \\
\hline
\end{tabular}
\end{center}
\end{example}

\subsection{Corpus Perspectives}
The elegance of the corpus representation lays in its compactness and 
universality that, however, yields for many diverse possibilities of the 
underlying data analysis. The analysis are enabled by the process of so 
called matricisation of the corpus tensor $\mathbf{C}$. Essentially, 
matricisation is a process of representing a higher-order tensor using a 
2-dimensional matrix perspective. This is done by fixing one tensor index as 
one matrix dimension and generating all possible combinations of the other 
tensor indices within the remaining matrix dimension. In the following we 
illustrate the process on the 
corpus tensor from Example~\ref{ex:corpus}. 

\begin{example}\label{ex:matricisation}
\small
When fixing the subjects ($A_l$ set members) of the corpus tensor from 
Example~\ref{ex:corpus}, one will get the following matricised perspective
(the rows and columns with zero values are omitted):
\begin{center}
\tiny
\begin{tabular}{|c||c|c|c|c|c|}
\hline
$s$/$\langle p,o \rangle$ & $\langle D,p.\rangle$ & $\langle T,d. \rangle$ & 
$\langle T,m. \rangle$ & $\langle I,j. \rangle$ & $\langle I,e.d. \rangle$ \\
\hline\hline
{\it p.d.} & $1/7$ & $2/7$ & $0$ & $0$ & $0$ \\ 
\hline
{\it g.} & $1/7$ & $0$ & $0$ & $0$ & $0$\\ 
\hline
{\it i.t.d.} & $0$ & $0$ & $1/7$ & $1/7$ & $1/7$ \\
\hline
\end{tabular}$\;$.
\end{center}
The row and column index abbreviations correspond to Example~\ref{ex:source}. 
One can see that the transformation is lossless, as the original tensor can be 
easily reconstructed from the matrix by appropriate re-grouping of the indices.
\end{example}

The corpus tensor matricisations correspond to vector spaces consisting of 
elements defined by particular rows of the matrix perspectives. Each row 
vector has a name (the corresponding matrix row index) and a set of features
(the matrix column indices). The features represent the distributional 
attributes of the entity associated with the vector's name -- the contexts
aggregated across the whole corpus. This can be used for various types of 
analysis and for inference of more complex semantic features emerging within
the simple statements extracted from the source data. In the following 
sections, we describe two particular types of analysis that are relevant to 
the motivating use case of this paper:
\begin{inparaenum}[(1)]
  \item computation of related (semantically close) terms;
  \item mining of conjunctive IF-THEN rules from the data.
\end{inparaenum}

\subsection{Computing Related Terms}
By comparing the row vectors in corpus tensor matricisations, one essentially 
compares the meaning of the corresponding label terms, as it is emerging from 
the underlying data. For exploring the matricised perspectives, one can 
use linear algebra methods that have been proven to work by countless 
successful applications to vector space analysis in the last couple of 
decades~\cite{Salton:1975:VSM:361219.361220,Deerwester90indexingby,manninetal2008irbook}. 
Large feature spaces can be reliably reduced to more manageable and less noisy
number of dimensions by techniques like singular value decomposition or 
random indexing (see \url{http://en.wikipedia.org/wiki/Dimension_reduction}). 
After the (optional) dimensionality reduction, the perspective vectors can be 
compared in a well-founded manner by measures like cosine similarity (see 
\url{http://en.wikipedia.org/wiki/Cosine_similarity}), as illustrated in 
Example~\ref{ex:analysis}.

\begin{example}\label{ex:analysis}
\small
Let us add one more matrix perspective to the $s$/$\langle p,o \rangle$ one 
provided in Example~\ref{ex:matricisation}. It represents the distributional 
features of right arguments (based on the contexts of relation terms and 
left arguments they tend to co-occur with in the corpus):
\begin{center}
\tiny
\begin{tabular}{|c||c|c|c|c|c|}
\hline
$o$/$\langle p,s \rangle$ & $\langle D,p.d.\rangle$ & $\langle T,p.d. \rangle$ 
& $\langle D,g. \rangle$ & $\langle T,i.t.d. \rangle$ & $\langle I,i.t.d. 
\rangle$ \\
\hline\hline
{\it p.} & $1/7$ & $0$ & $1/7$ & $0$ & $0$ \\ 
\hline
{\it d.} & $0$ & $2/7$ & $0$ & $0$ & $0$\\ 
\hline
{\it m.} & $0$ & $0$ & $0$ & $1/7$ & $0$ \\
\hline
{\it j.} & $0$ & $0$ & $0$ & $0$ & $1/7$ \\
\hline
{\it e.d.} & $0$ & $0$ & $0$ & $0$ & $1/7$ \\
\hline
\end{tabular}$\;$.
\end{center}

The vector spaces induced by the matrix perspectives $s$/$\langle p,o \rangle$
and $o$/$\langle p,s \rangle$ can be used for finding similar terms by 
comparing their corresponding vectors. Using the cosine vector similarity,
one finds that
$
sim_{s/\langle p,o \rangle}(p.d.,g.) = 
\frac{(1/7) \cdot (1/7)}{\sqrt{(1/7)^2+(2/7)^2}\sqrt{(1/7)^2}} \doteq 0.2972
$ 
and 
$
sim_{o/\langle p,s \rangle}(j.,e.d.) = 
\frac{(1/7) \cdot (1/7)}{\sqrt{(1/7)^2}\sqrt{(1/7)^2}} = 1.
$
These are the only non-zero similarities among the terms present in the 
corpus. This corresponds to the intuitive interpretation of the data 
represented by the initial statements from Example~\ref{ex:source}. Protein 
domains and genes seem to be different from proteins, yet protein domain is a 
type of domain and gene is not, therefore they share some similarities but are 
not completely equal according to the data. Juxtamembranes and extracellular 
domains are both places where internal tandem duplications can occur, and no 
other information is available, so they can be deemed equal (until more data 
comes). 
\end{example}

It can be easily seen how the computation of related terms is relevant to the 
annotation use case that has motivated the paper. By computing MeSH terms 
related to the content of an article (i.e., terms that have been extracted 
from it), one can get annotations that are semantically related to the article 
even if they are not present in it and/or linked to it in any explicit way. 

\subsection{Rule Mining}
Another type of emergent semantic pattern we can infer from the matricised 
corpus perspectives are IF-THEN rules. Rules are useful for our motivating 
use case due to their applicability to extension of the basic article 
annotations -- once we know that an article has annotations that conform to a 
rule's antecedent, we can also add annotations present in the rule consequent.

To simplify the presentation, let us consider conjunctive IF-THEN rules of 
type $(?x,l_1,r_1) \wedge (?x,l_2,r_2) 
\wedge \dots \wedge (?x,l_k,r_k) \rightarrow (?x,l_{k+1},r_{k+1}) \wedge 
(?x,l_n,r_n)$ in the following, where $?x$ is a variable and $l_i, r_i, i \in 
\{1,\dots,n\}$ are concrete relation and (right) argument terms. An example of 
such rule is $(?x,\mathrm{type\;of}, domain) \rightarrow 
(?x,\mathrm{different}$ $\mathrm{from},protein)$, which says that everything 
that is a 
type of domain is not a protein. The rule mining consists of two steps:
\begin{inparaenum}[(1)]
  \item using the matrix perspective $\langle p,o \rangle$/$s$ for finding 
  candidate sets of $\langle l_i, r_i \rangle)$ tuples that can form rules;
  \item using the matrix perspective $s$/$\langle p,o \rangle$ for pruning 
  the generated rules based on their confidence. 
\end{inparaenum}
Note that other types of single-variable conjunctive IF-THEN rules (i.e., the 
ones with variable occurring in the second or third position of the rule 
statements) can be computed in the same way, only using different 
perspectives.

The first step corresponds to finding all frequent itemsets in a database
as described in~\cite{Agrawal:1993:MAR:170036.170072}. The row vectors of the 
$\langle p,o \rangle$/$s$ matrix are essentially the `items' -- features of 
the rules, i.e., the concrete $(l_i, r_i), i \in \{1,\dots,n\}$ tuples. By 
grouping close vectors, we can discover related features that may possibly 
form rules. Perhaps a simplest way of doing this is $k$-means clustering based 
on Euclidean distance~\cite{hartigan1979} applied to the $\langle p,o 
\rangle$/$s$ matrix. The $k$ parameter is set so that the sizes of the 
generated clusters correspond to the desired maximum number of statements 
present in a rule. In practice, we recommend to apply dimensionality reduction
to the columns of the matrix. 
This makes the clustering faster, while also leading to noise reduction and 
better representation of the features' meaning in the sense 
of~\cite{Deerwester90indexingby}. The described approach effectively 
replaces the process of finding frequent itemsets 
in~\cite{Agrawal:1993:MAR:170036.170072}. Using our distributional 
representation, we find promising `itemsets' not via support in discrete
data transactions, but 
by exploiting their continuous latent semantics. 

The second step involves pruning of the previously generated rules using 
measures of {\it support} ($supp$) and {\it confidence} ($conf$). Only rules 
with sufficiently high confidence are kept as a result of the mining process. 
The measures are computed on a matrix that is a transpose of the one used for 
generating the rules ($s$/$\langle p,o \rangle$ in case of the discussed type 
of rules). We keep the original dimensions of the matrix this time, so that we 
can check for the confidence of the rules using the actual data without any 
transformations. 

We base the rule pruning on the definitions of support and confidence provided
in~\cite{Agrawal:1993:MAR:170036.170072}, however, we generalise the support 
so that we can fully exploit the power of our distributional representation.
The classic definition of $supp(X)$ for an itemset (set of features to form 
rule statements) is the relative frequency of rows in the data that contain 
the items in $X$.  
This is due to the 
fact that the data representation in classical rule mining is crisp -- the rows
(transactions) contain only zeros and ones that indicate the lack and presence
of an item in a transaction, respectively. Our data representation is more 
general -- zeros in the matrix still mean lack of an item in the given row,
however, the actual presence of items is represented in a more fluid way by 
real-valued weights. Therefore we define the generalised support 
as a function 
$supp: 2^F \rightarrow \mathbb{R}$, where $F$ is a set of rule features (i.e., 
the $\langle p, o \rangle$ column labels of the corpus perspective matrix on 
which the rules are being tested -- $s$/$\langle p,o \rangle$ for the type of 
rules discussed above). The support of a feature set $X$ on a perspective 
matrix $M$ is computed as $supp(X) = \frac{1}{||M||} \sum_{i \in 
I_X}\frac{\sqrt{\sum_{j \in X} m_{i,j}^2}}{|X|}$. $I_X$ is a set of all row 
indices of the matrix $M$ where all the features from $X$ are present (i.e.,
have a non-zero value), and $m_{i,j}$ is an element of the matrix $M$ with 
indices $i,j$. $||M||$ is a matrix norm (i.e., `size') defined as $||M|| = 
\sum_{i \in I}\frac{\sum_{j \in J}m_{i,j}}{|\{m_{i,k}|k \in J \wedge m_{i,k} 
\neq 0\}|}$, where $I,J$ are sets of row and column indices of $M$, 
respectively. The confidence of a rule $X \rightarrow Y$ is then computed as 
defined in~\cite{Agrawal:1993:MAR:170036.170072}, i.e., $conf(X \rightarrow 
Y) = \frac{supp(X \cup Y)}{supp(X)}$, only using the generalised support. 
The process of rule mining 
is further illustrated in Example~\ref{ex:rules} in the end of this section.

The proposed definition of support essentially computes weighted relative 
frequency of the input feature set $X$ in the matrix rows. Only rows that 
contain all features contribute to the absolute frequency count. The actual 
contribution is computed as a normalised Euclidean size of the row vector 
restricted only to the column indices from $X$. The normalising factor is the 
size of the feature set (this to make the support value independent on the 
size of $X$). The absolute weighted frequency of the feature set is then 
divided by $||M||$ to get the relative frequency (analogically to the classical
definition of support). $||M||$ reflects the size of the real-valued data set 
as a sum of all weights in the matrix $M$, normalised by the number of non-zero
elements per each row 
(this makes also the norm independent on the size of all potential feature 
sets). One can easily check that if the values in the matrix $M$ are just 
zeros and ones as in the traditional data representation used
by~\cite{Agrawal:1993:MAR:170036.170072}, our support becomes the classical 
one. 

\begin{example}\label{ex:rules}
\small
Building on the previous examples, the `training' matrix for the rule mining 
is a transpose of the one given in Example~\ref{ex:matricisation}: 
\begin{center}
\tiny
\begin{tabular}{|c||c|c|c|}
\hline
$\langle p,o \rangle$/$s$ & {\it p.d.} & {\it g.} & {\it i.t.d.} \\
\hline\hline
$\langle D,p.\rangle$ & $1/7$ & $1/7$ & $0$ \\
\hline
$\langle T,d. \rangle$ & $2/7$ & $0$ & $0$ \\
\hline
$\langle T,m. \rangle$ & $0$ & $0$ & $1/7$ \\
\hline
$\langle I,j. \rangle$ & $0$ & $0$ & $1/7$ \\
\hline
$\langle I,e.d. \rangle$ & $0$ & $0$ & $1/7$ \\
\hline
\end{tabular}$\;$.
\end{center}
No dimension reduction is applied due to simplicity of the example. The 
testing matrix is the original one, i.e., $s$/$\langle p,o \rangle$ 
from Example~\ref{ex:matricisation}. The Euclidean distance between any 
two of the last three vectors in the `training' matrix $\langle p,o 
\rangle$/$s$ is $0$. The distance between the first two vectors 
is $d_{1,2} = \sqrt{(1/7-2/7)^2 + (1/7-0)^2} = \frac{\sqrt{2}}{7}$. The 
distances between the first and second and any of the last three vectors are 
$d_{1,3\mathrm{-}5} = \sqrt{(1/7-0)^2 + (1/7-0)^2 + (0-1/7)^2} = 
\frac{\sqrt{3}}{7}$ and $d_{2,3\mathrm{-}5} = \sqrt{(2/7-0)^2 + (0-1/7)^2} = 
\frac{\sqrt{5}}{7}$, respectively. The minimum-distance grouping of the 
vectors into clusters containing at least two elements is thus as follows: 
$G_1: \{\langle D,p.\rangle, \langle T,d. \rangle\}, G_2: \{\langle
T,m. \rangle, \langle I,j. \rangle, \langle I,e.d. \rangle\}$. 

Let us abbreviate the rule statements corresponding to the `training' matrix 
above as follows: $s_1: (?x,D,p.), s_2: (?x,T,d.), s_3: (?x,T,m.), s_4: 
(?x,I,j.), s_5: (?x,I,e.d.)$. Then the groups $G_1, G_2$ generate these
$14$ rules: $R_1\mathrm{-}R_2: s_i \rightarrow s_j, i,j \in \{1,2\}$, 
$R_3\mathrm{-}R_5: s_i \rightarrow s_j \wedge s_k, i,j,k \in \{3,4,5\}$, 
$R_6\mathrm{-}R_8: s_i \wedge s_j \rightarrow s_k, i,j,k \in \{3,4,5\}$, 
$R_9\mathrm{-}R_{14}: s_i \rightarrow s_j, i,j \in \{3,4,5\}$. 

The `testing' matrix is $s$/$\langle p,o \rangle$ (see 
Example~\ref{ex:matricisation}) and its size is $0.5$. 
Corresponding supports of the relevant sets of rule statements are: 
$supp(\{s_1\}) = supp(\{s_2\}) = 
\frac{4}{7}, supp(\{s_3\}) = supp(\{s_4\}) = supp(\{s_5\}) = \frac{2}{7}, 
supp(\{s_1,s_2\}) = \frac{\sqrt{5}}{7}, supp(\{s_3,s_4\}) = 
supp(\{s_3,s_5\}) = supp(\{s_4,s_5\}) = \frac{\sqrt{2}}{7}, 
supp(\{s_3,s_4,s_5\}) = \frac{2\sqrt{3}}{21}$. Thus the confidences of the 
rules are: $conf(R_1) = conf(R_2) = \frac{\sqrt{5}}{4}, 
conf(R_3) = conf(R_4) = conf(R_5) = \frac{\sqrt{3}}{3},
conf(R_6) = conf(R_7) = conf(R_8) = \frac{\sqrt{6}}{3},
conf(R_9) = conf(R_{10}) = \dots = conf(R_{14}) = \frac{\sqrt{2}}{2}$. When 
setting the confidence threshold to $0.5$, the rules $R_1\mathrm{-}R_5$ are 
discarded. 
\end{example}


\section{Automated Document Annotation}\label{sec:practice}

This section illustrates the practical potential of the general framework 
introduced so far. First we describe its application to unsupervised 
annotation of biomedical articles with terms from the MeSH thesaurus. Then we 
present the evaluation of our approach and discuss the results obtained.

\subsection{Data and Method}

As a corpus of documents for annotation, we employed $2,003$ articles from
the PubMed repository (\url{http://www.ncbi.nlm.nih.gov/pubmed/}) that had 
their fulltexts 
available from PubMed Central 
(\url{http://www.ncbi.nlm.nih.gov/pmc/}). The articles were selected so 
that for each article present, the corpus also contained corresponding related
articles as offered by the PubMed's related articles 
service~\cite{citeulike:1845865}. This fact was important for the evaluation 
later on. 
For the article annotation, we used the MeSH 2011 version (obtained at 
\url{http://www.nlm.nih.gov/mesh/filelist.html}). 

We processed the data using the following high-level pipeline:
\begin{inparaenum}[(1)]
  \item extraction of statements from the articles and from MeSH;
  \item incorporation of the extracted statements into 
  two separate knowledge bases for PubMed articles and for MeSH thesaurus;
  \item construction of basic MeSH annotation sets for each article;
  \item mining of rules from the MeSH knowledge base;
  \item rule-based extension of the basic annotation sets;
  \item evaluation of the initial and extended sets of annotations. 
\end{inparaenum}

In the extraction step, we were focusing on simple binary co-occurrence 
statements. We tokenized the article text into sentences, then applied part of 
speech tagging and shallow parsing in order to determine noun phrases. Any two 
noun phrases $NP_1,NP_2$ occurring in the same sentence formed a statement 
$(NP_1,R,NP_2)$, where $R$ stands (here and in the following) for a 
$related\_to$ relationship expressing a general relatedness between the left 
and right arguments. Any synonyms of MeSH terms in the statements were 
converted to the corresponding preferred MeSH headings in order to lexically 
unify the data. $1,379,235$ statements were generated from the 
$2,003$ articles this way. From the MeSH data set, we generated $(T_1,R,T_2)$ 
statements for all terms (i.e., headings) $T_1, T_2$ such that they were
parent, child or sibling of each other in the MeSH hierarchy, which led to
$41,632$ statements. Note that for both data sets, we considered the $R$ 
relation symmetric, which effectively made the $s/\langle p,o \rangle$ and 
$o/\langle s,p \rangle$ perspectives equivalent in the consequent steps. 

The adopted model of co-occurrence limited to a single general relationship $R$
may seem to be restrictive, however, we chose to do so to be able to link the 
semantics of the data extracted from articles with the semantics of MeSH in 
the most general sense applicable. 
Apart of that, \cite{Novacek2011iswc} suggest that in settings similar to ours,
such `flattened' semantics 
actually perform better than a model with multiple relations. 

The second step in the experimental pipeline was incorporation of the
extracted statements into knowledge bases (i.e., the source, corpus and 
perspective structures described in Section~\ref{sec:theory}). The 
incorporation was done in the same way for both PubMed and MeSH data. 
The source ($\mathbf{G}$) values were set to $1$ for all elements 
$g_{s,p,o,d}$ such that the statement $(s,p,o)$ occurred in the document $d$; 
all other values were $0$. To get the corpus ($\mathbf{C}$) tensor values 
$c_{s,p,o}$, we multiplied the frequency of the $(s,p,o)$ triples (i.e., 
$\sum_{d \in P}g_{s,p,o,d}$) by the point-wise mutual information score of the 
$(s,o)$ tuple (see 
\url{http://en.wikipedia.org/wiki/Pointwise_mutual_information}).

The annotations for each article $d$ were computed using the article knowledge 
base as follows. First we constructed a set $TF = \{(t,f_d(t))|t \in d\}$, 
where $t$ are all terms extracted from $d$ and $f_d(t)$ is the absolute 
frequency 
of the term $t$ in $d$. For each $(t,f_d(t))$ tuple from $TF$, we computed 
another set $REL_t = \{(t^{\prime},f_d(t) \cdot 
sim_{s/\langle p,o \rangle}(t,t^{\prime}))|
sim_{s/\langle p,o \rangle}(t,t^{\prime}) > 0\}$. Rephrased in prose, the 
$REL_t$ sets contained tuples of all terms similar to $t$ and the actual 
similarities multiplied by the $f_d(t)$ frequency (more frequent terms should 
generally produce terms with higher relatedness value). The 
$sim_{s/\langle p,o \rangle}$ similarity function was defined as in 
Example~\ref{ex:analysis}. Eventually, we collated the particular 
term relatedness values across the whole 
document $d$ into an overall relatedness $rel(t^{\prime}) = 
\frac{1}{W}\sum_{w \in W_{t^{\prime}}}w$, where $W_{t^{\prime}} = \{r|
(t^{\prime},r) \in \bigcup_{t \in d} REL_t\}$ and $W$ is a sum of all the
relatedness values occurring in the $\bigcup_{t \in d} REL_t$ union. The final 
output of this step for each document $d$ was a set of all related terms 
$t^{\prime}$ such that $t^{\prime}$ is in MeSH. The $rel(t^{\prime})$ values 
were used for ranking the set of MeSH annotations and taking only the top ones 
if necessary. 

The rule mining part of the experimental pipeline was 
executed iteratively with different random initialisations of the clusters 
until no new rules were added in at least $10$ most recent iterations. 
We obtained $33,384$ rules with confidence at least $0.5$ this way. The rules 
were then used for extending the basic article annotation sets as follows. 
Let us assume an article $d$ has annotations $\{t_1, t_2, \dots, t_n\}$. 
Then for any rule $(?x,R,e_1) \wedge (?x,R,e_2) \wedge \dots \wedge (?x,R,e_k) 
\rightarrow (?x,R,e_{k+1}) \wedge (?x,R,e_{k+2}) \wedge \dots \wedge 
(?x,R,e_m)$ such that $\{t_1, t_2, \dots, t_n\} \subseteq 
\{e_1, e_2, \dots, e_k\}$, we used the consequent set $\{e_{k+1}, e_{k+2}, 
\dots, e_m\}$ as extended annotations for the article $d$. The relatedness
measure of the extensions $e$ was computed as $\frac{1}{W}\sum_{w \in C_e}w$, 
where $C_e$ is a set of confidences of all rules that contributed with the 
extension $e$, and $W$ is a sum of all such confidences across all extensions 
computed. Similarly to the basic annotation sets, the relatedness of the 
extensions was used for their ranking and possible restriction to top-scoring 
ones. 

Note that the data we have been working with, as well as the library and 
scripts we have implemented for the experiment, are available for reference at
\url{http://dl.dropbox.com/u/21379226/aaai2012_761.zip}. 

\subsection{Evaluation and Discussion}

To evaluate the annotation sets produced in the experimental pipeline, we used 
two methods. 
  Firstly, we measured precision and recall of the basic and extended 
  annotation sets based on their comparison with manually provided MeSH 
  annotations of the corresponding articles (available through the PubMed's 
  Entrez API). For each article, we computed average precision, precision and 
  recall~\cite{manninetal2008irbook} of all computed annotations and also of 
  top $h$ ones, where $h$ is the number of human annotations for the given 
  article.
  
  The second evaluation method focused on the utility of the computed
  annotations, namely in the task of finding related articles. We used a 
  standard vector space model~\cite{Salton:1975:VSM:361219.361220} for 
  determining the relatedness of documents, where features were formed by the 
  sets of computed or manually assigned article annotations. For each document,
  we computed different sets of related documents (based on the human 
  annotations and on the basic/extended ones generated by our framework). 
  To determine their precision and recall, the computed sets were compared 
  to corresponding sets of related articles provided by the dedicated PubMed 
  service\footnote{The service is based on algorithms described 
  in~\cite{citeulike:1845865}. This is obviously less desirable than a gold 
  standard designed solely by human experts, however, no such gold standard 
  was readily available for all PubMed articles we processed and we lacked the 
  manpower to create it ourselves. In this situation, we considered the state 
  of the art service currently endorsed by the PubMed staff and millions of 
  users as a reasonable alternative to a hand-crafted gold standard.}. 
  Similarly to the evaluation of annotations themselves, we measured average
  precision, precision and recall of all and of top $h$ related articles 
  computed, where $h$ was the number of related articles 
  in the gold standard. 

The results of the evaluation are summarised in Tables~\ref{tab:eval1} 
and~\ref{tab:eval2}. The mean average precision (MAP), precision and 
recall 
lines in the tables were computed
as an arithmetic mean across the particular values for all $2,003$ articles in 
the experimental corpus. The F-score ($F_1$ in particular) was computed from 
the mean precision/recall values. The columns in the tables correspond to the 
types of the article annotation sets described above. BASE, EXT. refer to the 
basic and extended annotations, while ALL, TOP refer to complete and top-$h$ 
only annotation sets. Note that we did not include the EXT./TOP annotations 
into the result summaries, since they were performing significantly worse than 
the other ones in most of the measured categories.
\begin{table}[ht]
\center
\tiny
\begin{tabular}{|c||c|c|c|}
\hline
            & \multicolumn{2}{c|}{\bf BASE} & {\bf EXT.} \\
            & {\it ALL} & {\it TOP}     & {\it ALL} \\ 
\hline\hline
{\bf MAP}   & 5.5 & {\bf 5.5}  &  5.3 \\
\hline
{\bf prec.} & 14.9 & {\bf 16.4}  &  9.3 \\
\hline
{\bf rec.}  & 12.2 & 11.7  &  {\bf 12.7} \\
\hline
{\bf F-sc.} & 13.4 & {\bf 13.7}  &  10.7 \\
\hline
\end{tabular}
\caption{Evaluation results (article annotation)}
\label{tab:eval1}
\end{table}



The comparison with the manually curated MeSH annotations in 
Table~\ref{tab:eval1} does not look particularly impressive, with highest 
precision and recall values of $16.4\%$ and $12.7\%$, respectively. On the 
other hand, the automatically computed annotations performed much better than 
the `manual' ones when using them as features for finding related articles. 
\begin{table}[ht]
\center
\tiny
\begin{tabular}{|c||c|c|c|c|c|c|c|c|c|c|c|c|c|c|}
\hline
                & \multicolumn{2}{c|}{\bf BASE-ALL} & 
                  \multicolumn{2}{c|}{\bf BASE-TOP} &
                  \multicolumn{2}{c|}{\bf EXT-ALL} &
                  \multicolumn{2}{c|}{\bf HUMAN} \\
                & {\it ALL} & {\it TOP}         & {\it ALL} & {\it TOP} &
                  {\it ALL} & {\it TOP}         & {\it ALL} & {\it TOP} \\
\hline\hline
{\bf MAP}   & 21.6 & 21.7 & 14.6 & 22.3 & {\bf 36.9} & 36.9 & 21.1 & 21.9 \\
\hline
{\bf prec.} & 64.7 & 65 & 49.9 & 79.2 & {\bf 91.1} & 91.1 & 45.3 & 46.9 \\
\hline
{\bf rec.}  & 52 & 51.8 & 58.3 & 36.3 & 41.8 & 41.8 & {\bf 68.5} & 67.9 \\
\hline
{\bf F-sc.} & {\bf 57.7} & {\bf 57.7} & 53.8 & 49.8 & 57.3 & 57.3 & 54.5 & 55.5 \\
\hline
\end{tabular}
\caption{Evaluation results (annotation utility)}
\label{tab:eval2}
\end{table}
As can be seen in Table~\ref{tab:eval2}, there is a substantial improvement
namely regarding precision and overall F-score. The only measure where the 
manually curated annotations perform slightly (ca. $1.1$-times) better than 
the next-best automated method is recall. Especially notable is the difference 
in precision -- the extended annotations achieve more than $91\%$, which is 
about two-times better than the human ones. 


The results we obtain may have several interpretations. We believe that one of 
the more plausible ones is related to the nature of the manually provided MeSH
annotations. As mentioned for instance in~\cite{neveol2009jbi}, the goal of
PubMed annotators is to provide best MeSH `tags' for the purpose of indexing 
in digital library collections. Thus they are motivated to select annotations 
that better discriminate papers
from each other. This may, however, be rather detrimental when the task is 
to identify related papers using the annotations, as features used for 
identifying relatedness (i.e., similarity) are often dual to the features used
for discrimination of entities~\cite{tversky77similarity}. This reasoning can 
in turn explain why our automatically computed article annotations, apparently 
very different from the manually curated ones, perform significantly better 
when used as features for finding related articles. The better performance 
(especially in case of the precision of extended annotations) may indicate 
that the automatically computed annotations are selected in a more 
fine-grained manner and from a more varied `vocabulary' than the ones provided
by human annotators, who can hardly grasp the scale of all the hypothetically 
available annotations (in addition to having different motivations as 
mentioned before). This is not to say that either kind of 
annotations is worse than the other, it much rather means that they simply 
serve slightly different purposes. 

To conclude the discussion, 
we believe that despite of
the low performance of our approach in terms of comparison with manually
curated MeSH annotations, we can still offer potentially very beneficial 
results (especially in case of annotations augmented by emergent rules). This 
holds particularly for use cases where the annotations are 
supposed to be produced in a scalable and economical way in order to determine 
similarities between articles. Examples of such use cases include not only
identification of related documents, but also question answering or automated 
linking of publications and supplementary data (e.g., biomedical data in the 
RDF format provided at \url{http://linkedlifedata.com/sources}, which we can 
easily incorporate as implied 
by~\cite{Novacek2011iswc}).


\section{Conclusions and Future Work}\label{sec:conclusion}

We presented an approach to acquisition of complex knowledge patterns emerging 
within simple statements extracted from textual data. The distinctive 
features of our approach are unification of the principles of emergent and 
distributional semantics, and a novel method for mining rules from the 
proposed distributional representation. To demonstrate the practical relevance 
of our work, we applied it to annotation of PubMed articles with terms from 
the MeSH thesaurus. After discussing our results, we identified areas where 
our approach can likely bring most benefits to users.

In future, we will explore more use cases and investigate other types of 
knowledge patterns (e.g., emergent formation of new candidate concepts and 
taxonomical relations to be 
recommended for inclusion into the MeSH thesaurus). Regarding the presented
use case, we intend to look into possible combinations of our approach and 
relevant state of the art (namely the ranking-based methods 
like~\cite{huangetal2011jamia} or~\cite{ruch2006bioinf}). This is also related 
to deeper evaluation of our work that would utilise the state 
of the art approaches as a base-line (currently we were not able to do so 
comprehensively enough due to lack of publicly available and applicable 
implementations). Eventually, we want to perform a qualitative evaluation of 
the annotations produced by our system with an assistance of domain experts.


\bibliographystyle{plain}
\bibliography{bib_data}

\end{document}